\documentclass[10pt,a4paper]{article}

\usepackage{gensymb}
\usepackage{makeidx} 
\usepackage{graphics}
\usepackage{epsfig}
\usepackage{graphicx}
\usepackage{verbatim}
\usepackage{amsfonts}
\usepackage{times}
\usepackage{amstext}
\usepackage{amsmath}
\usepackage{amssymb}
\usepackage{amsthm}

\usepackage{algorithm}
\usepackage{url}
\usepackage{multicol}
\usepackage{multirow}
\usepackage[table]{xcolor}
\usepackage{booktabs}
\usepackage{lineno}
\usepackage{color,soul}
\soulregister\cite7
\soulregister\citep7
\soulregister\ref7

\newcommand\numberthis{\addtocounter{equation}{1}\tag{\theequation}}

\usepackage[margin=1.5in]{geometry}

\def\ANNp{ANN\ }

\def\ANNs{ANNs}
\def\ANNsp{ANNs } 

\def\ASNNs{SNNs}
\def\ASNNsp{SNNs\ }

\def\ASNN{SNN}
\def\ASNNp{SNN\ }

\title{Efficient Computation in Adaptive Artificial Spiking Neural Networks}

\author{Davide Zambrano$^{1}$, Roeland Nusselder$^1$, H. Steven Scholte$^2$ \& Sander Boht\'e$^1$}

\newcommand{\PLH}{{\mkern-2mu\times\mkern-2mu}}
\begin{document}

\maketitle

\paragraph{Abstract. }
Artificial Neural Networks (ANNs) are bio-inspired models of neural computation that have proven highly effective.
Still, ANNs lack a natural notion of time, and neural units in ANNs exchange analog values in a frame-based manner, a computationally and energetically inefficient form of communication. 
This contrasts sharply with biological neurons that communicate sparingly and efficiently using binary spikes. 
While artificial Spiking Neural Networks (SNNs) can be constructed by replacing the units of an 
ANN with spiking neurons \cite{Cao2015ijcv,diehlIJCNN2015}, the current performance is far from that of deep ANNs on hard benchmarks and these SNNs use much higher firing rates compared to their biological counterparts, limiting their efficiency.
Here we show how spiking neurons that employ an efficient form of neural coding can be used to construct \ASNNsp that match high-performance ANNs and exceed state-of-the-art in SNNs on important benchmarks, while requiring much lower average firing rates. 
For this, we use spike-time coding based on the firing rate limiting adaptation phenomenon observed in biological spiking neurons. This phenomenon can be captured in adapting spiking neuron models, for which we derive the effective transfer function. Neural units in ANNs trained with this transfer function can be substituted directly with adaptive spiking neurons, and the resulting Adaptive \ASNNsp (Ad\ASNNs) can carry out inference in deep neural networks using up to an order of magnitude fewer spikes compared to previous \ASNNs. 
Adaptive spike-time coding additionally allows for the dynamic control of neural coding precision: we show how a simple model of arousal in Ad\ASNNsp further halves the average required firing rate and this notion naturally extends to other forms of attention. Ad\ASNNsp thus hold promise as a novel and efficient model for neural computation that naturally fits to temporally continuous and asynchronous applications.
  \paragraph{Introduction}

While the currently best-performing \ASNNsp use high firing rates (on average hundreds of Hertz) to cover the dynamic range of corresponding analog neurons \cite{Cao2015ijcv,diehlIJCNN2015}, in biology, real neurons use on average 1-5Hz\cite{attwell2001energy} and sensory neurons are known to adaptively control the number of spikes that are used to efficiently cover large dynamic ranges \cite{fairhall2001efficiency}. This adaptive behaviour can be captured with fast spike-triggered adaptation in Leaky-Integrate-and-Fire neuron models, or corresponding Spike Response Models (SRMs)\cite{GerKis02,Bohte:2012tf,Pozzorini:2013bj} including the Adaptive Spiking Neuron models (ASN)\cite{Bohte:2012tf}. 
ASNs can implement adaptive-spike coding as a neural coding scheme that maps analogue values to sequences of spikes, where the thresholding mechanism carries out an online analog-to-digital conversion of the analog signal computed in the neuron unit. 

\begin{figure}[ht!]
\centering
  \includegraphics[width=10.5cm]{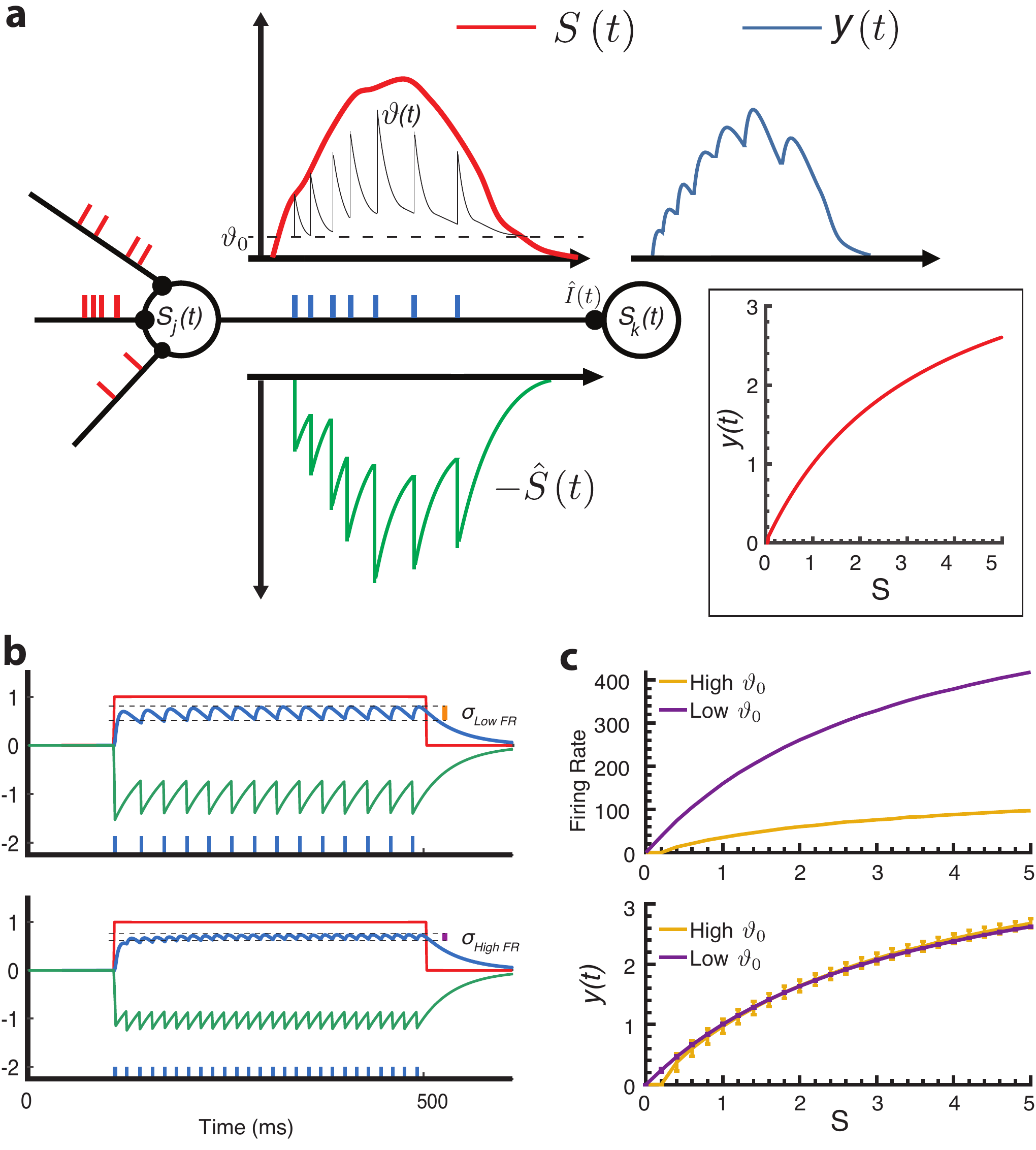}
  \caption{{\bf a}, Adaptive spike-time coding. Spikes are generated when the difference between the activation $S(t)$ and the refractory response $\hat{S}(t)$ exceeds $\frac{1}{2}\vartheta(t)$. Emitted spikes contribute a sum of PSPs to the target neuron's potential. Inset: effective transfer function.  {\bf b}, Example of encoding a step-function activation: the resultant normalised postsynaptic contribution $y(t)$  (blue line) is plotted for two different values of the resting threshold $\vartheta_0 = [0.5,0.1]$. The default neural coding precision ($\sigma_{\text{Low FR}}$) increases
   for a lower threshold $\vartheta_0$ corresponding to a higher firing rate ($\sigma_{\text{High FR}}$). {\bf c}, Top: two values of $\vartheta_0$ result in two different firing rate curves; bottom: the effective transfer function for two values of $\vartheta_0$, the same approximated value can be represented with different precisions (standard-deviation across the average), and thus different firing rates, by controlling spike height $h$.}
  \label{fig:fig1}
\end{figure}

Adaptive spike-time coding is illustrated in Fig. \ref{fig:fig1}a: expressed as an SRM, the input to a neuron $j$ consists of an external input $V_{\text{inj},j}(t)$, and a series of spikes from input neurons $i$ impinging at times $t_i$ each contributing a postsynaptic potential (PSP) weighted by synaptic efficacy $w_{ij}$. The PSP is modelled as a normalised kernel $\kappa(t-t_i)$ multiplied by the height of the spike $h$. These contributions result in the neuron's activation $S_j$(t): 
\begin{align}
S_j(t) & = V_{\text{inj},j}(t) + \sum_i \sum_{t_i} w_{ij} h \kappa(t-t_i).
\end{align}
The activation $S_j(t)$ corresponds to the membrane potential of a spiking neuron absent spiking. 
A spike emitted by neuron $j$ at time $t_j$ resets the membrane potential by subtracting a scaled refractory kernel $\vartheta(t_j)\eta(t)$, this kernel is added to the total refractory response $\hat{S}_j(t)$ that is computed as the sum of scaled refractory kernels; $\hat{S}_j(t)$ thus approximates the rectified activation $[S_j(t)]^+$. 
A spike is emitted when the the membrane potential -- the difference between $S_j(t)$ and $\hat{S}_j(t)$ -- exceeds half the threshold $\vartheta(t)$ (as in \cite{Yoon:hv}, the threshold $\vartheta$ is redefined for convenience). Spike-triggered adaptation is incorporated into the model by multiplicatively increasing the variable threshold $\vartheta(t)$ at the time of spiking with a decaying kernel $\gamma(t)$:
\begin{equation}
	\vartheta_j(t) = \vartheta_{0} + \sum\limits_{t_j} m_{f} \vartheta(t_j) \gamma(t-t_j),
\end{equation}
where $\vartheta_{0}$ is the resting threshold and the multiplicative parameter $m_{f}$ controls the speed of the firing rate adaptation. 
This adaptive spiking mechanism effectively maps an activation $S_j$ to a normalised average contribution $y(S_j)$ to the next neuron's activation $S_k$ as a rectified half-sigmoid-like transfer function (Fig. \ref{fig:fig1}a, inset): 
\begin{align}
y_j  = f(S_j)  = \left< \sum_{t_j}  \kappa(t-t_j) \right>,
\end{align}
We derive an analytical expression for the shape of the transfer function $f(S)$ to map spiking neurons to analog neural units (see Methods). The use of exponentially decaying kernels for $\eta(t), \gamma(t)$ and $\kappa(t)$ allows the neuron model to be computed with simple dynamical systems where each impinging spike adds an impulse response function of height $h$ multiplied by the weight associated with the connection.

The speed of adaptation, $m_f$, and the spike height $h$ together control the precision of the spike-based neural coding,
where neural coding precision is measured as the standard-deviation of $y(t)$ around the mean response to a fixed input. 
 As illustrated in Fig. \ref{fig:fig1}b, a same-but-more-precise spike-time encoding can be realised by changing the adaptation parameters $m_f, \vartheta_0$ to increase the firing rate for a given stimulus intensity, while simultaneously reducing the impact of spikes on target neurons by decreasing $h$. An ASN can thus map different stimulus-to-firing-rate curves (Fig. \ref{fig:fig1}c, top) to the same transfer function but with different neural coding precision (Fig. \ref{fig:fig1}c, bottom).

\begin{figure}[ht!]
\centering
  \includegraphics[width=14cm]{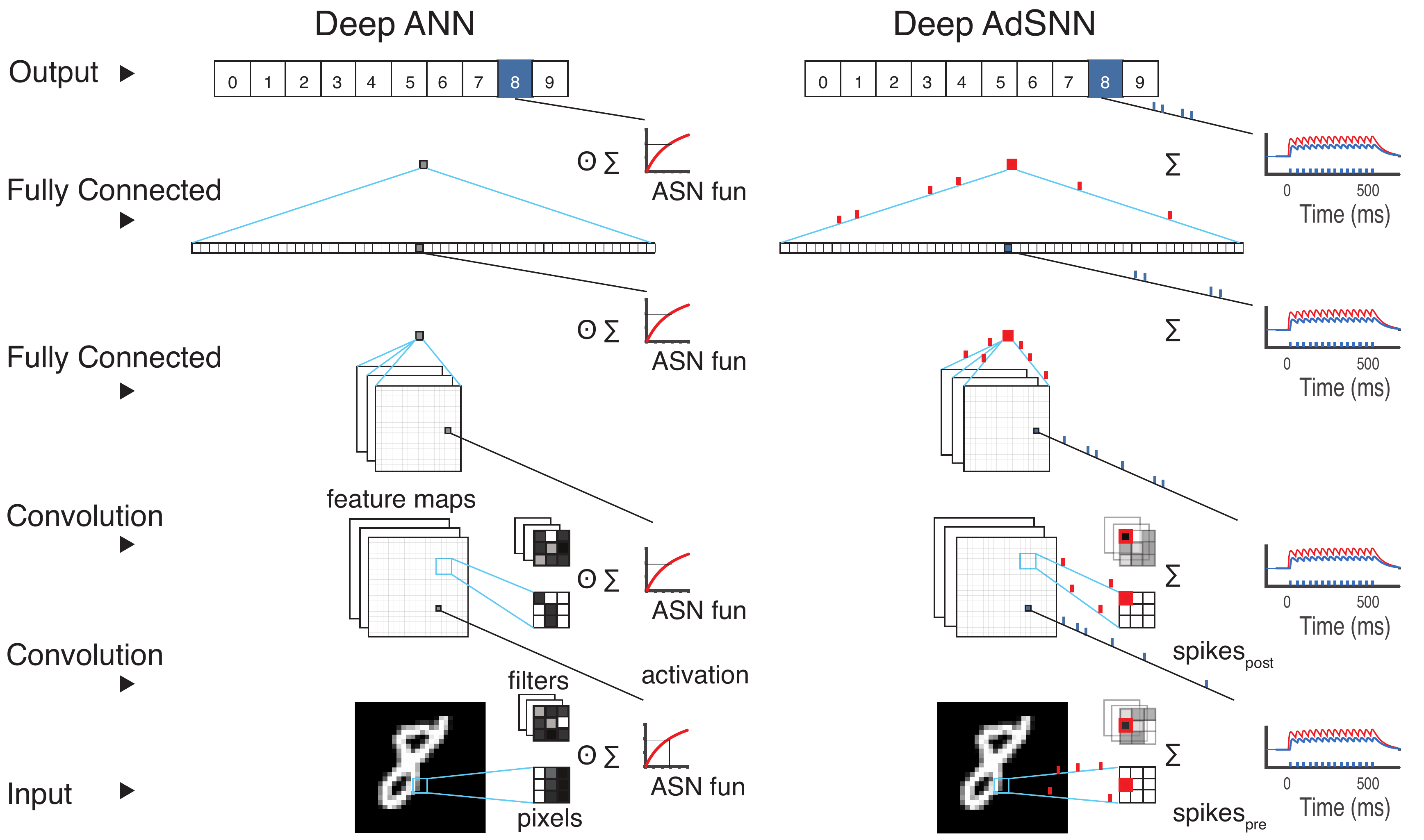}
  \caption{Adaptive Spiking Neural Network conversion schematic. Left. During training of a deep ANN, the output of a convolutional or fully connected layer is passed through the ASN transfer function. In ANNs, every layer performs a series of weighted sums of the inputs as each analog input is multiplied by its analog weight. Right. For classification, the analog ANN units are converted in ASNs to obtain an SNN. All the layers perform similarly, but now binary spikes are conveyed across the network and corresponding weights are simply added to the next ASN. 
  }
  \label{fig:network}
\end{figure}

We construct adaptive SNNs -- Ad\ASNNs -- comprised of ASN neurons using adaptive spike-coding similar to the approach pioneered in \cite{diehlIJCNN2015}. First, \ANNsp are constructed with analog neural units that use the derived half-sigmoid-like transfer function, both for fully connected feed-forward \ANNsp 
and for various deep convolutional neural network architectures. We train these \ANNsp for standard benchmarks of increasing difficulty (SONAR, IRIS, MNIST, CIFAR-10/100, and the ImageNet Large-Scale Visual Recognition Challenge (ILSVRC 2012) benchmarks). Corresponding Ad\ASNNsp 
are then obtained by replacing the \ANNs' analog units with ASNs (illustrated in Fig. \ref{fig:network}). 

\begin{table}[t]

  \caption{\small Performance($\%$), Matching Firing Rate (FR) (Hz) and Matching Time (MT) (ms). (*): for CIFAR-10, \cite{Hunsberger2016-hg} report performance of $83.54$ @ 143Hz. Current (Poisson) SNN performance and firing rate is compared against trained ANN and AdSNN performance. State of the art is denoted with bold font; no current SNN state of the art exists.}
  \label{tab:lowFR}
  \centering
  
  \begin{tabular*}{\textwidth}{llr|r|lrr}
 	\cmidrule{1-7}  
	\multirow{2}{*}{DataSet} & \multicolumn{2}{c|}{Current SNN} & ANNs & \multicolumn{3}{c}{AdSNNs} \\
	\cmidrule{2-7}  
 	& Perf. & FR  & Perf. & Perf. & FR  & MT \\

 	\cmidrule{1-7}  
 	
    IRIS   & - & - & $98.67$ & $\mathbf {98.67\pm2.8e^{-15}}$ & $51$ &  $269$ \\
    SONAR  & - & - & $89.42$ & $\mathbf {89.89\pm1.1}$ & $26$ &  $119$ \\
    MNIST & $99.12^{\text{\cite{Hunsberger2016-hg}}}$ & $\mathbf {9.47}^{\text{\cite{Hunsberger2016-hg}}}$
    & $99.56$ & $\mathbf {99.56\pm4.6e^{-3}}$ & $67$ &  $291$ \\
    CIFAR-10& $87.86^{\text{\cite{rueckauer2016theory}}*}$ & -  
    & $89.90$  & $\mathbf {89.86\pm4.2e^{-2}}$ & $68$ & $372$ \\
    CIFAR-100 &  $55.13^{\text{\cite{Hunsberger2016-hg}}}$ & - 
    &  $64.53$ & $\mathbf {64.21\pm2.4e^{-2}}$ & $59$ & $507$ \\
    ILSVRC-2012 & $51.8^{\text{\cite{Hunsberger2016-hg}}}$ & $93^{\text{\cite{Hunsberger2016-hg}}}$ & $62.98$ & $\mathbf {62.97\pm4.8e^{-2}}$ & $66$ & $347$ \\
     	\cmidrule{1-7}  
  \end{tabular*}
\end{table}

\begin{table}[t]

  \caption{\small High precision AdSNNs (from Table 1) Matching Time (MT) (ms) to achieve current (Poisson) SNN performance.  Performance($\%$), Firing Rate (FR) (Hz) and Matching Time (MT) (ms) are also provided for low precision AdSNNs that have similar current (Poisson) SNN performance. }
  \label{tab:matcSoA}
  \centering
  
  \begin{tabular*}{\textwidth}{l|lr|c|lrr}
	\cmidrule{1-7}  
	\multirow{2}{*}{DataSet} & \multicolumn{2}{c|}{Current SNN} & \begin{minipage}{1.25cm}AdSNNs Table1 \end{minipage}&\multicolumn{3}{c}{Low FR AdSNNs} \\
	\cmidrule{2-7}  
 	& Perf. & FR  & MT & Perf. & FR  & MT\\
	\cmidrule{1-7}  
       MNIST & $99.12^{\text{\cite{Hunsberger2016-hg}}}$ & $ {9.47}^{\text{\cite{Hunsberger2016-hg}}}$
     & $131$ &$ {99.12\pm4.14e^{-2}}$ & $\mathbf{10}$ &  $209$ \\
    CIFAR-10& $87.86^{\text{\cite{rueckauer2016theory}}*}$ & -  
     & $276$ & $ {88.52\pm2.9}$ & $\mathbf{22}$ & $304$ \\
    CIFAR-100 &  $55.13^{\text{\cite{Hunsberger2016-hg}}}$ & - 
     & $286$ & $ {58.08\pm9.5}$ & $\mathbf{16}$ & $328$ \\
    ILSVRC-2012 & $51.8^{\text{\cite{Hunsberger2016-hg}}}$ & $93^{\text{\cite{Hunsberger2016-hg}}}$  & $256$ & $ {53.77\pm5.4}$ & $\mathbf{12}$ & $338$ \\

    	\cmidrule{1-7}  
  \end{tabular*}
\end{table}

\begin{table}[t]

  \caption{\small AdSNNs versus Arousal AdSNNs. Performance($\%$), Matching Firing Rate (FR) (Hz) and Matching Time (MT) (ms). }
  \label{tab:arousal}
  \centering
  
  \begin{tabular*}{\textwidth}{l|lrr|lrr}
	\cmidrule{1-7}  
	\multirow{2}{*}{DataSet} & \multicolumn{3}{c|}{AdSNNs} & \multicolumn{3}{c}{Arousal AdSNNs}\\
	\cmidrule{2-7}  
 	& Perf. & FR  & MT & Perf. & FR  & MT\\
	\cmidrule{1-7}  
    IRIS   & $\mathbf {98.67\pm2.8e^{-15}}$ & $51$ &  $269$ & $98.67\pm0$ & $\mathbf {18}$ & $484$\\
    SONAR  & $\mathbf {89.89\pm1.1}$ & $26$ &  $119$ & $90.25\pm0.3$ & $\mathbf { 15}$ & $393$\\
    MNIST & $\mathbf {99.56\pm4.6e^{-3}}$ & $67$ &  $291$ & $99.56\pm0$ & $12$ & $498$\\
    CIFAR-10 &  $\mathbf {89.86\pm4.2e^{-2}}$ & $68$ & $372$& $89.88\pm4.5e^{-2}$ & $\mathbf {34}$& $546$\\
    CIFAR-100 & $\mathbf {64.21\pm2.4e^{-2}}$ & $59$ & $507$ & $64.57\pm0.12$ & $\mathbf {50}$& $691$\\
    ILSVRC-2012 & $\mathbf {62.97\pm4.8e^{-2}}$ & $66$ & $347$ & $63.02\pm0.1$ &$\mathbf {42}$ & $575$ \\
    	\cmidrule{1-7}  
  \end{tabular*}
\end{table}

For suitable choices of adaptation parameters (Table SI1), the Ad\ASNNsp exactly match performance to the original \ANNsp as measured on the test set (Table \ref{tab:lowFR}). Since we trained high-performance \ANNs, the Ad\ASNNsp exceed previous state-of-the-art \ASNNp performance on all benchmarks while requiring substantially lower average firing-rates, in the range of 24-68 Hz. Note that on some benchmarks, the Ad\ASNNsp exceed the \ANNsp performance, presumably because the Ad\ASNNsp compute an average from sampled neural activity\cite{Hunsberger2016-hg} that correctly separates some additional inputs.
As any \ASNN, the time-based communication in Ad\ASNNsp incurs latency, measured as the time required between onset of the stimulus and the time when the output neurons are able to classify at the level of the network's analog counterpart. For Ad\ASNNs, this latency (MT) is of order 300ms, and mainly depends on the PSP decay time (50ms here); faster decay times result in lower latency, at the expense of increased firing rates (see Fig. SI1a). However, state-of-the-art accuracies are already reached after about 200ms (Table \ref{tab:matcSoA}).

\begin{figure}[ht!]
\centering
  \includegraphics[width=12cm]{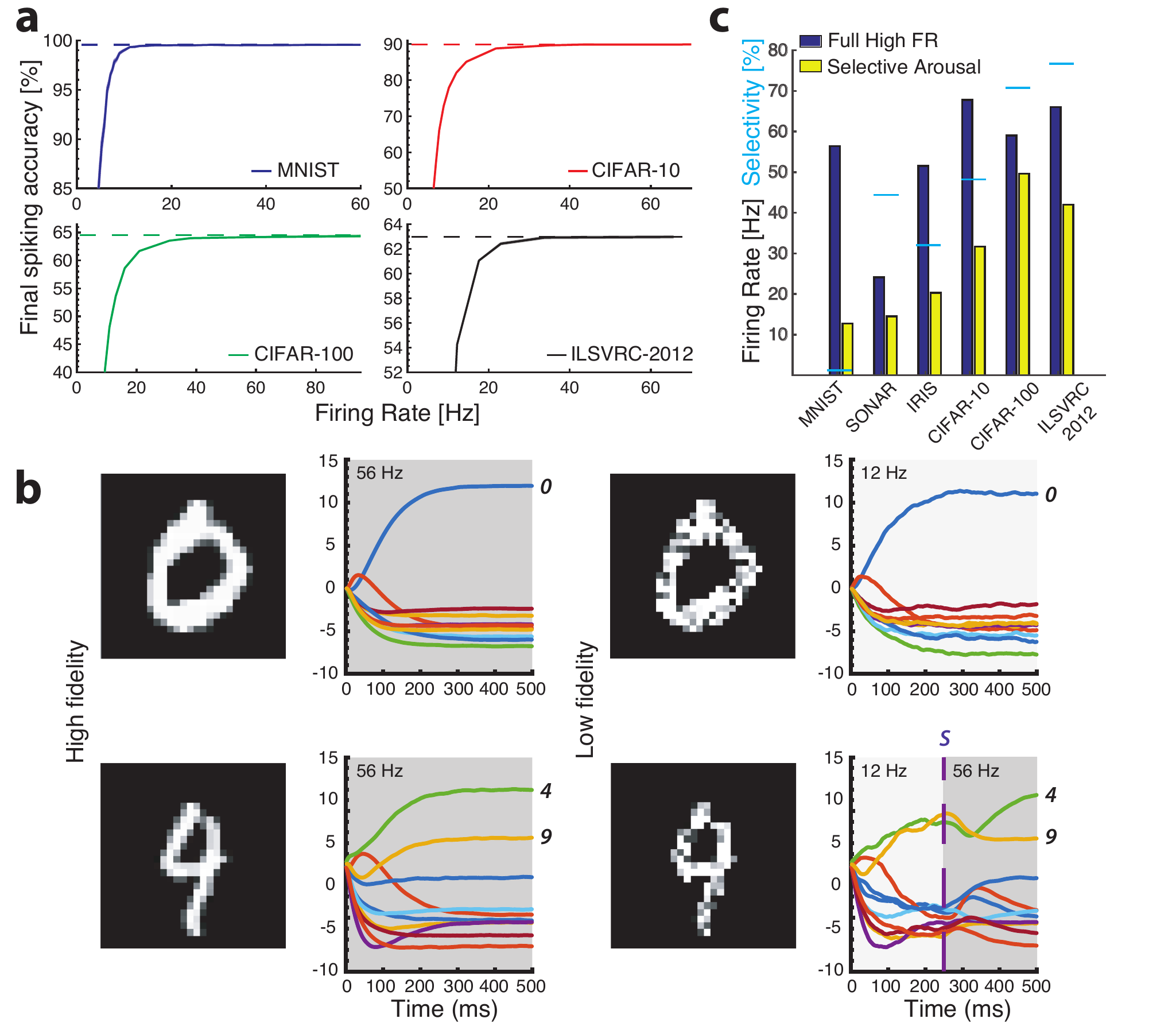}
  \caption{{\bf a}, accuracy on the test sets of four datasets (MNIST, CIFAR-10 and CIFAR-100, ILSVRC-2012) as a function of average firing rate: lower coding precision decreases accuracy. {\bf b}, Classifying with attention. Ease of classification is measured as the distance between the internal value $S$ of the winning output neuron and the 
second highest output neuron (line-plots). Top row: easy example that is correctly classified both at low precision (right) and high precision (left). Bottom row: ambiguous samples can be disambiguated by applying arousal to increase precision in the network. {\bf c}, Efficiency improvement when Arousal is applied to hard-to-classify images only: the same classification accuracy is reached using a significantly lower average firing rate over the test set. Cyan bars designate percentage of inputs selected by the Arousal criterion.}
  \label{fig:fig3}
\end{figure}

We further find that Ad\ASNNsp exhibit a gradual and graceful performance degradation when the neural coding precision is decreased, by changing the ASN adaptation parameters such that the firing rate is lowered while increasing $h$ (Fig. \ref{fig:fig3}a): performance equal to previously reported state-of-the-art can be reached with even lower firing rates (10-22Hz, Table \ref{tab:matcSoA}). Increasing the PSP decay time further lowers the required firing rate to achieve Ad\ASNNp performance matching ANNs (Fig. SI1b), at the expense of increased latency (Fig. SI1a)\footnote{We excluded \cite{esser2016convolutional} as there a binary neural network is simulated without notion of time while requiring many more binary neurons as compared to similarly performing ANNs.}.

The tuneable relationship between firing rate and neural coding precision can be exploited to further increase efficiency by selectively manipulating this trade off as a particular form of attention. It is well known that for stable sensory inputs attention in the brain manifests as enhanced firing in affected neurons \cite{Roelfsema:1998hz}. One purported effect of this mechanism is to improve neural coding precision on demand, for instance in specific locations, for a brief amount of time, and only if needed \cite{saproo2010spatial,friston2010free}. Such attention would allow the brain to process information at a low default precision when possible and increase firing rate only when necessary, potentially saving a large amount of energy.

We implement attention in the form of arousal affecting all neurons in the network simultaneously. Arousal is engaged selectively based on classification uncertainty: the neural coding precision is increased from a low base level only for samples deemed uncertain, as illustrated in Fig. \ref{fig:fig3}b. Uncertain inputs are identified by accumulating the two highest valued outputs for $50$ms after a network-dependent fixed waiting time (dashed vertical line in Fig. \ref{fig:fig3}b). Arousal is engaged only if the averaged difference between these two outputs does not exceed a hard threshold as determined from the training set; engaging arousal causes a brief deterioration of classification accuracy before quickly settling to higher performance (Fig. SI2). 
Using this simple model of attentional modulation, the number of spikes required for overall classification is effectively halved (Fig. \ref{fig:fig3}c), while latency increases as the selected inputs require additional time for classification (see Table \ref{tab:arousal}). 
The uncertainty based arousal is also engaged more or less frequently depending on the accuracy of the model (blue markers in Fig. \ref{fig:fig3}c), and the benefit is thus greatest for networks with the highest absolute accuracy.

\paragraph{Discussion. } 
A number of recent studies have suggested that spiking neurons implement an efficient analog-to-digital conversion similar to the mechanisms proposed here \cite{deneve2011,Bohte:2012tf,Yoon:hv}. While population coding is a popular concept to explain how pools of spiking neurons can approximate analog signals with arbitrary precision \cite{Deneve2016-ku}, the results presented here show that firstly, the required neural coding precision in many deep neural networks can be satisfied with a single and plausible spiking neuron model at reasonable firing rates; and secondly, neural coding precision can further be increased or decreased by manipulating the firing rate inversely with  a form of global synaptic efficacy modulation. This provides an alternative explanation for the observed attentional modulation of firing rates, and more detailed location-based or object-based attention algorithms can be studied to increase neural efficiency further. 

As presented, the half-sigmoid-like derived transfer function holds for isomorphic spikes that can be communicated with a binary number.
A rectified linear (ReLU) transfer function can be constructed by scaling the impact of individual spikes on postsynaptic targets with the presynaptic adaptation magnitude at the time of spiking. Using such a transfer function slightly improves performance and speeds up convergence, at the expense of effectively communicating an analog value with each spike (not shown). 
From a biological perspective, such neural communication would require a tight coupling between neural adaptation and phenomena like synaptic facilitation and depression \cite{abbott2004synaptic}, which at present has not been examined in this context. From a computer science perspective, the efficiency penalty in terms of bandwidth may be limited as spike-based neuromorphic simulators like SpiNNaker already use sizable addressing bits for each spike \cite{furber2013overview}; the computationally simple addition of spikes to the target neuron however is replaced by a conventional multiply-add operation. 

Ad\ASNNsp explicitly use the time-dimension for communication and implicitly exploit temporal correlations in signals for sparse spike-time coding; in contrast, ANNs applied to temporal problem domains sequentially and synchronously sample their inputs in a time-stepped manner, recomputing the network for each successive timestep. The actual extraction of computational efficiency from sparsely active \ASNNsp is a separate challenge. 
Sparse activity and computationally cheap connection updates are accompanied by a more complex and state-based neuron model that is updated more frequently. Networks with a high fan-in fan-out architecture, like the brain, benefit most from this trade-off; current deep learning architectures in contrast are characterised by a low degree of fan-in fan-out, except for the last layers which are typically fully connected. Hybrid analog/spiking neural network approaches may be most efficient for the implementation of these architectures. Additionally, like other state-based neural networks, and in contrast to feedforward \ANNp architectures, networks of adapting spiking neurons require per-neuron local memory to store state information such as potential and adaptation values. The availability of sufficient local memory is thus necessary to best extract efficiency from sparse spiking activity. Since current GPU-based deep learning accelerators are lacking in this regard, at least for the large state-based neural networks considered, neuromorphic hardware seems the most suitable approach for the implementation of large \ASNNs.

\section*{Methods}
\paragraph{Adaptive Spiking Neurons.}

In the ASN, the kernel $\kappa(t)$ is computed as the convolution of a spike-triggered postsynaptic current (PSC) with a filter $\phi$, decaying exponentially respectively with time constants $\tau_{\phi}$ and $\tau_{\beta}$; and the input signal $V_{\text{inj},j}(t)$ is similarly computed from a current injection $I_{\text{inj},j}(t)$. The adaptation kernel $\gamma(t)$ decays with time-constant $\tau_{\gamma}$.
 
The Ad\ASNNs's are created by converting standard Deep Neural Networks\cite{diehlIJCNN2015} trained with a mathematically derived transfer function $f(S)$ of the ASN (full derivation in SI), defined as the function that maps the activation $S$ to the average post-synaptic contribution. This has the form: 
\begin{equation*}
\label{eq:activation_function}
f(S) = \max\left(0,\frac{h}{\exp\left(\frac{c_1 \cdot S + c_2}{c_3 \cdot S + c_4}\right) - 1} - c_0 + h/2\right),
\end{equation*}
where,
\begin{align*}
c_1 &= 2\cdot m_f \cdot \tau_{\gamma}^2, \\
c_2 &= 2\cdot \vartheta_0 \cdot \tau_{\eta} \cdot \tau_{\gamma}, \\
c_3 &= \tau_{\gamma}\cdot (m_f \cdot \tau_{\gamma} + 2\cdot(m_f+1)\cdot \tau_{\eta}), \\
c_4 &= \vartheta_0 \cdot \tau_{\eta} \cdot (\tau_{\gamma} + \tau_{\eta}), \\ 
c_0 &=\frac{h}{\exp\left(\frac{c1 \cdot \vartheta_0/2 + c2}{c3 \cdot \vartheta_0/2 + c4}\right) - 1},
\end{align*} 
are constants computed from the neuron parameters setting, and $h$ defines the spike size. 
Here, by normalising $f(S)$ to $1$ when $S = 1$, $h$ becomes a scaling factor for the network's trained weights, allowing communication with binary spikes. 
\paragraph{Adaptive Spiking Neural Networks (Ad\ASNNs).}
Analog units using $f(S)$ as their transfer function, AANs, in trained \ANNsp can be replaced directly and without modification with ASNs. In the presented results, the adaptation kernel decays with $ \tau_{\gamma} = 15 ms$, the membrane filter with $\tau_{\phi}=5ms$, the refractory response with $\tau_{\eta} = 50 ms$ and the PSP with $\tau_{\beta} = 50 ms$, all roughly corresponding to match biological neurons, and $\vartheta_{0} = m_{f}$. Batch Normalization (BN)\cite{ioffe2015batch} is used to avoid  the vanishing gradient problem\cite{hochreiter1998vanishing} for saturating transfer functions like half-sigmoids and to improve the network training and regularisation. After training, the BN layers are removed and integrated into the weights' computation \cite{rueckauer2016theory}. 
A BN-AAN layer is also used as a first layer in all the networks to convert the inputs into spikes. When converting, biases are added to the post-synaptic activation. 
Max and Average Pooling layers are converted by merging them into the next ASN-layer: the layer activation $S$ is computed from incoming spikes, then the pooling operator is applied and the ASN-layer computes spikes as output.
The last ASN layer acts as a smoothed read-out layer with $\tau_{\phi}=50 ms$, where spikes are converted into analog values for classification. The classification is performed as in the ANN network, usually using SoftMax: at every time-step $t$ the output with highest value is considered the result of the classification. 
\paragraph{\ANNp training.}
We trained \ANNp with AANs on widely used datasets: for feedforward ANNs, IRIS and SONAR; and for deep convolutional ANNs: MNIST, CIFAR-10, CIFAR-100 and ILSVRC-2012. 
All the \ANNsp are trained using Keras\footnote{\url{https://keras.io/}} with Tensorflow\footnote{\url{https://www.tensorflow.org/}} as its backend. 
We used categorical cross-entropy as a loss function with Adam\cite{kingma2014adam} as the optimiser, except for ILSVRC-2012 where we used Stochastic Gradient Decent with Nesterov (learning rate $ = 1e-3$, decay $=1e-4$ and momentum $=0.9$). Consistent with the aim of converting high performance ANNs into AdSNNs, for each dataset, we selected the model at the training epoch where it performed best on the test set. 

We trained a $[4-60-60-3]$ feedforward ANN on the IRIS dataset: IRIS is a classical non-linearly separable toy dataset containing $3$ classes -- $3$ types of plants -- with $50$ instances each, to be classified from $4$ input attributes. Similarly, for the SONAR dataset\cite{Gorman:1988dx} we used a $[60-50-50-2]$ ANN to classify $208$ entries of sonar signals divided in $60$ energy measurements in a particular frequency band in two classes: metal cylinder or simple rocks. We trained both ANNs for $800$ epochs and obtained competitive performance. 

The deep convolutional ANNs are trained on standard image classification problems with incremental difficulty. The simplest is the MNIST dataset\cite{Lecun:1998hy}, where $28\PLH28$ images of handwritten digits have to be classified. We used a convolutional ANNs composed of $[28 \PLH 28-c64 \PLH 3-m2-2\PLH(c128 \PLH 3-c)- m2-d256-d50-10]$, where $cN \PLH M$ is a convolutional layer with $N$ feature maps and a kernel size of $M \PLH M$, $mP$ is a max pooling layer with kernel size $P \PLH P$, and $dK$ is a dense layer with $K$ neurons. Images are pre-normalised between $0$ and $1$, and the convolutional ANN was trained for $50$ epochs. 

The CIFAR-10 and CIFAR-100 data sets\cite{krizhevsky2009learning} are harder benchmarks, where $32\PLH32$ colour images have to be classified in 10 or 100 categories respectively. We use a VGG-like architecture\cite{simonyan2014very} with $12$ layers: $[32 \PLH 32 - 2\PLH(c64 \PLH 3) - m2 - 2\PLH(c128 \PLH 3) - m2 - 3\PLH(c256 \PLH 3) - m2 - 3\PLH(c512 \PLH 3)- m2 - d512 - 10]$ for CIFAR-10 and $[32 \PLH 32 - 2\PLH(c64 \PLH 3) - m2 - 2\PLH(c128 \PLH 3) - m2 - 3\PLH(c256 \PLH 3) - m2 - 3\PLH(c1024 \PLH 3)- m2 - d1024 - 100]$ for CIFAR-100. Dropout\cite{srivastava2014dropout} was used in the non-pooling layers ($0.5$ in the top fully-connected layers, and $0.2$ for the first $500$ epochs and $0.4$ for the last $100$ in the others). Images were converted from RGB to YCbCr and then normalised between $0$ and $1$.

The ImageNet Large-Scale Visual Recognition Challenge (ILSVRC)\cite{ILSVRC15} is a large-scale image classification task with over 15 million labeled high-resolution images belonging to roughly $22,000$ categories. The 2012 task-1 challenge was used, a subset of ImageNet with about $1000$ images in each of $1000$ categories. We trained a ResNet-18 architecture in the Identity-mapping variant \cite{he2016deep} for 100 epochs and the top-1 error rate is reported. As in\cite{simonyan2014very}, we rescaled the images to a resolution of $256 \PLH 256$ pixels and then performed random cropping during training and centre cropping for testing.

\paragraph{Ad\ASNNp evaluation.}
The Ad\ASNNsp are evaluated in simulations with 1$ms$ timesteps, where inputs are persistently presented for $500ms$ (identical to the method used in \cite{diehlIJCNN2015}). The Firing Rate (FR) in Table \ref{tab:lowFR} is computed as the average number of spikes emitted by a neuron, for each image, in this time window. The time window is chosen such that all output neurons reach a stable value; we defined the Matching Time (MT) as the time to which $99\%$ of the maximum classification accuracy is reached for each simulation. From the MT to the end of the time interval, the standard deviation of the accuracy is computed to evaluate the stability of the network's response. Each dataset was evaluated for a range of $\vartheta_0,m_f$ values of $[0.03, 1.0]$ and the minimum firing rate needed to match the \ANNp performance is reported. All the Ad\ASNNsp simulations are run on MATLAB in a modified version of the MatConvNet framework\footnote{\url{http://www.vlfeat.org/matconvnet/}}. 

\paragraph{Arousal.}
For Arousal, we highlight uncertain inputs by increasing firing-rate and corresponding precision. The network is simulated with $\vartheta_0$ set to $\vartheta_{0-lp}$, the standard low-precision parameter; if the input is selected by the arousal mechanism, this parameter is set to high precision value: $\vartheta_{0-hp}$ (and $m_f$ is changed identically). 
Selection is determined by accumulating the winning and the 2nd-highest outputs for $50ms$ starting from a pre-defined $t_{sa}$ specific for each dataset. If the difference between these two outputs exceeds a threshold $\theta_A$, the input is not highlighted -- $\theta_A$ is estimated by observing those images that are not correctly classified when the precision is decreased on the training set. The Arousal method selects more images than needed: we defined Selectivity as the proportion of highlighted images  (Table SI1 ). In addition, $\theta_A$ increases linearly with the accumulation time interval as $\theta_A = p_{1}\cdot (t-t_{sa}) + p_{2}$, while Selectivity decreases exponentially. 
We report results for the minimum firing rate recorded for each dataset (Fig. \ref{fig:fig3}c), 
which is obtained at a specific $\vartheta_{0-lp}$: in fact, starting from very low precision leads to higher Selectivity, which in turn results in a higher average firing rate. The parameter $\vartheta_{0-hp}$ is chosen as the lowest precision needed to match the \ANNp performance. Table SI1 reports the values of Selectivity, $t_{sa}, \vartheta_{0-lp}, \vartheta_{0-hp}, p_{1}, p_{2}$ for each dataset. Note that, since deeper networks need more time to settle to the high precision level, we extended the simulation time for these networks (see Table \ref{tab:lowFR}).

\section*{Supplementary Information}
To convert a trained Artificial Neural Network (ANN) into an Adaptive Spiking Neural Network (AdSNN), the transfer function of the ANN units needs to match the behaviour of the Adaptive Spiking Neuron (ASN). The ASN transfer function is derived for the general case of $\tau_{\eta} \neq \tau_{\gamma}$ using an approximation of the ASN behaviour. 

\paragraph{Derivation of the ASN activation function}
We consider a spiking neuron with activation $S(t)$ that is constant over time, and the refractory response $\hat{S}(t)$ approximates $S(t)$ using a variable threshold $\vartheta(t)$. Whenever $S - \hat{S}(t) > 0.5 \cdot \vartheta(t)$, the neuron emits a spike of fixed height $h$ to the synapses connecting to the target neurons, and a value of $\vartheta(t_{f})$ is added to $\hat{S}$, with $t_{f}$ the time of the spike. At the same time, the threshold is increased by $m_{f} \vartheta(t_{f})$. 
The post-synaptic current (PSC) in the target neuron is then given by $I(t)$, which is convolved with the membrane filter $\phi(t)$ to obtain the contribution to the post-synaptic potential; a normalized exponential filter $\phi(t)$ with short time constants $\tau_\phi$ will just smooth the high-frequency components of $I(t)$.  We derive the transfer function that maps the activation $S$ to the PSC $I$ of the target neuron.
We recall the ASN model here, elaborating the SRM to include the current-to-potential filtering:
\begin{align}
	\label{eq:PSC}
	&\text{PSC:} & I(t) &= \sum_{i}\sum_{t_{s}^{i}} w_{i} \exp\left(\frac{t_{s}^{i} - t}{\tau_{\beta}}\right), \\
	\label{eq:Sinput}
	&\text{activation:} & S(t) &= (\phi \ast I)(t), \\
	\label{eq:theta}
	&\text{threshold:} & \vartheta(t) &= \vartheta_{0} + \sum\limits_{t_{s}} m_{f} \vartheta(t_{s}) \exp\left(\frac{t_{s} - t}{\tau_{\gamma}}\right), \\
	\label{eq:Shat}
	&\text{refractory response:} & \hat{S}(t) &= \sum\limits_{t_{s}} \vartheta(t_{s}) \exp\left(\frac{t_{s} - t}{\tau_{\eta}}\right),
\end{align}

where $t_{s}^{i}$ denotes the timing of incoming spikes that the neuron receives and $t_{s}$ the timing of outgoing spikes. 

Since the variables of the ASN decay exponentially, they converge asymptotically. For a given fixed size current injection, we consider a neuron that has stabilised around an equilibrium, that is $\hat{S}(t)$ and $\vartheta(t)$ at the time of a spike always reach the same values. Let these values be denoted as $\hat{S}_{l}$ and $\vartheta_{l}$ respectively. Then, $\vartheta(t_{f}) = \vartheta_{l}$ and $\hat{S}(t_{f}) = \hat{S}_{l}$ for all $t_{f}$. The PSC $I(t)$ also always declines to the same value, $I_{l}$, before it receives a new spike. Setting $t=0$ for the last time that there was a spike, we can rewrite our ASN equations, Equations (\ref{eq:PSC}), (\ref{eq:Sinput}), (\ref{eq:theta}) and (\ref{eq:Shat}), for $\tau_{\beta} = \tau_{\eta}$ and $0 < t < t_{f}$ to:
\begin{align*}
    & \hat{S}(t) = \hat{S}_{l} e^{-\frac{t}{\tau_{\eta}}} + \vartheta_{l} e^{-\frac{t}{\tau_{\eta}}},\\
    & \vartheta(t) = \vartheta_{0} + (\vartheta_{l} - \vartheta_{0}) e^{-\frac{t}{\tau_{\gamma}}} + m_{f} \vartheta_{l} e^{-\frac{t}{\tau_{\gamma}}},\\
    & I(t) = I_{l} e^{-\frac{t}{\tau_{\eta}}} + h e^{-\frac{t}{\tau_{\eta}}}.
\end{align*}
The transfer function $f(S)$ of the ASN is a function of the value of $S$; $f(S)$ should be a bit larger than $I_{l}$ since that is the lowest value of $I(t)$, and we are interested in the average value of $I(t)$ between two spikes: $f(S) = I_{average}$.

Since we are in a stable situation, the time between each spike is fixed; we define this time as $t_{e}$. Thus, if the last spike occurred at $t=0$, the next spike should happen at $t = t_{e}$. This implies that $\hat{S}(t), \vartheta(t)$ and $I(t)$ at $t = t_{e}$ must have reached their minimal values $\hat{S}_{l}, \vartheta_{l}$ and $I_{l}$ respectively.

To obtain the activation function $f(S)$, we solve the following set of equations:
\begin{align*}
    & \hat{S}(t_{e}) = \hat{S}_{l},\\
    & \vartheta(t_{e}) = \vartheta_{l},\\
    & I(t_{e}) = I_{l},
\end{align*}
and by noting that the neuron only emits a spike when $S - \hat{S}(t) > \frac{1}{2} \vartheta$, we also have:
\begin{align*}
    & S - \hat{S}_{l} = \frac{1}{2}  \vartheta_{l}.
\end{align*}
We first notice:
\begin{align}
\label{eq:Is}
    \frac{h e^{-\frac{t_{e}}{\tau_{\eta}}}}{1 - e^{-\frac{t_{e}}{\tau_{\eta}}}} = I_{l}.
\end{align}
We now want an expression for $\vartheta_{l}$:
\begin{align*}
    & \vartheta_{0} + (\vartheta_{l} - \vartheta_{0}) e^{-\frac{t_{e}}{\tau_{\gamma}}} + m_{f} \vartheta_{l} e^{-\frac{t_{e}}{\tau_{\gamma}}} = \vartheta_{l},\\
    & \vartheta_{0} - \vartheta_{0} e^{-\frac{t_{e}}{\tau_{\gamma}}} = \vartheta_{l} - m_{f} \vartheta_{l} e^{-\frac{t_{e}}{\tau_{\gamma}}} - \vartheta_{l} e^{-\frac{t_{e}}{\tau_{\gamma}}}.
\end{align*}
We can rewrite this to:
\begin{align}
\label{eq:thetas}
    \vartheta_{0} \frac{1 - e^{-\frac{t_{e}}{\tau_{\gamma}}}}{1 - (m_{f} + 1) e^{-\frac{t_{e}}{\tau_{\gamma}}}} = \vartheta_{l}.
\end{align}
Using equations $S - \hat{S}_{l} = \frac{1}{2}  \vartheta_{l}$ and $\hat{S}(t_{e}) = \hat{S}_{l}$, we get:
\begin{align*}
    & (S - \frac{1}{2}  \vartheta_{l}) e^{-\frac{t_{e}}{\tau_{\eta}}} + \vartheta_{l} e^{-\frac{t_{e}}{\tau_{\eta}}} = S -\frac{1}{2} \vartheta_{l},\\
    & e^{-\frac{t_{e}}{\tau_{\eta}}}(2 S + \vartheta_{l}) = 2 S - \vartheta_{l}.
\end{align*}
Inserting Equation \ref{eq:thetas} gives:
\begin{align*}
    e^{-\frac{t_{e}}{\tau_{\eta}}} \left( 2 S + \vartheta_{0} \frac{1 - e^{-\frac{t_{e}}{\tau_{\gamma}}}}{1 - (m_{f} + 1) e^{-\frac{t_{e}}{\tau_{\gamma}}}} \right) = 2 S - \vartheta_{0} \frac{1 - e^{-\frac{t_{e}}{\tau_{\gamma}}}}{1 - (m_{f} + 1) e^{-\frac{t_{e}}{\tau_{\gamma}}}}.
\end{align*}
This can be rewritten to:
\begin{align*}
    e^{-\frac{t_{e}}{\tau_{\eta}}} \left( 2 S (1 - (m_{f} + 1) e^{-\frac{t_{e}}{\tau_{\gamma}}}) + \vartheta_{0} (1 - e^{-\frac{t_{e}}{\tau_{\gamma}}}) \right) &= 2 S (1 - (m_{f} + 1) e^{-\frac{t_{e}}{\tau_{\gamma}}})\\
    &- \vartheta_{0} (1 - e^{-\frac{t_{e}}{\tau_{\gamma}}}),\\
    (2 S + \vartheta_{0}) e^{-\frac{t_{e}}{\tau_{\eta}}} - (2 S (m_{f} + 1) + \vartheta_{0}) e^{-\frac{t_{e}}{\tau_{\gamma}}} e^{-\frac{t_{e}}{\tau_{\eta}}} &= 2 S - \vartheta_{0}\\
    &- 2 S (m_{f} + 1) e^{-\frac{t_{e}}{\tau_{\gamma}}}\\
    &+ \vartheta_{0} e^{-\frac{t_{e}}{\tau_{\gamma}}},\\
    (2 S + \vartheta_{0}) e^{-\frac{t_{e}}{\tau_{\eta}}} - (2 S (m_{f} + 1) + \vartheta_{0}) e^{-t_{e}(\frac{1}{\tau_{\gamma}} + \frac{1}{\tau_{\eta}})} &+ (2 S (m_{f} + 1) - \vartheta_{0}) e^{-\frac{t_{e}}{\tau_{\gamma}}}\\
    &= 2 S - \vartheta_{0}. \numberthis  \label{eq:endofboth}
\end{align*}

\paragraph{Approximation of the AAN activation function}
In the general case of $\tau_{\eta} \neq \tau_{\gamma}$, a (second order) Taylor series expansion can be used to approximate the exponential function:
\begin{align*}
    e^{x} \approx 1 + x + \frac{x^{2}}{2},
\end{align*}
for $x$ close to $0$. We can use this in our previous equation:
\begin{align*}
    (2 S + \vartheta_{0})&(1 - \frac{1}{\tau_{\eta}} t_{e} + \frac{1}{2 \tau_{\eta}^{2}} t_{e}^{2})\\
    &- (2 S (m_{f} + 1) + \vartheta_{0})(1 - (\frac{1}{\tau_{\gamma}} + \frac{1}{\tau_{\eta}}) t_{e} + \frac{1}{2}(\frac{1}{\tau_{\gamma}} + \frac{1}{\tau_{\eta}})^{2} t_{e}^{2})\\
    & + (2 S (m_{f} + 1) - \vartheta_{0})(1 - \frac{1}{\tau_{\gamma}} t_{e} + \frac{1}{2 \tau_{\gamma}^{2}} t_{e}^{2})\\
    &= 2 S - \vartheta_{0}.
\end{align*}
We need a few steps to isolate $t_{e}$:
\begin{align*}
    (2 S + \vartheta_{0})&( - \frac{1}{\tau_{\eta}} t_{e} + \frac{1}{2 \tau_{\eta}^{2}} t_{e}^{2})\\
    &- (2 S (m_{f} + 1) + \vartheta_{0})( - (\frac{1}{\tau_{\gamma}} + \frac{1}{\tau_{\eta}}) t_{e} + \frac{1}{2}(\frac{1}{\tau_{\gamma}} + \frac{1}{\tau_{\eta}})^{2} t_{e}^{2})\\
    & + (2 S (m_{f} + 1) - \vartheta_{0})( - \frac{1}{\tau_{\gamma}} t_{e} + \frac{1}{2 \tau_{\gamma}^{2}} t_{e}^{2}) = 0,\\
    (2 S + \vartheta_{0})&( - \frac{1}{\tau_{\eta}} + \frac{1}{2 \tau_{\eta}^{2}} t_{e})\\
    &- (2 S (m_{f} + 1) + \vartheta_{0})( - (\frac{1}{\tau_{\gamma}} + \frac{1}{\tau_{\eta}}) + \frac{1}{2}(\frac{1}{\tau_{\gamma}} + \frac{1}{\tau_{\eta}})^{2} t_{e})\\
    & + (2 S (m_{f} + 1) - \vartheta_{0})( - \frac{1}{\tau_{\gamma}} + \frac{1}{2 \tau_{\gamma}^{2}} t_{e}) = 0,
\end{align*}
\begin{align*}
    ((2 S + \vartheta_{0})\frac{1}{2 \tau_{\eta}^{2}} &- (2 S (m_{f} + 1) + \vartheta_{0})\frac{1}{2}(\frac{1}{\tau_{\gamma}} + \frac{1}{\tau_{\eta}})^{2}\\
    &+ (2 S (m_{f} + 1) - \vartheta_{0})\frac{1}{2 \tau_{\gamma}^{2}} )t_{e}\\
    & = (2 S + \vartheta_{0})\frac{1}{\tau_{\eta}} - (2 S (m_{f} + 1) + \vartheta_{0})(\frac{1}{\tau_{\gamma}} + \frac{1}{\tau_{\eta}})\\
    &+ (2 S (m_{f} + 1) - \vartheta_{0})\frac{1}{\tau_{\gamma}},
\end{align*}
\begin{align*}
    (- S \cdot m_{f}\frac{1}{\tau_{\eta}^{2}} &- (2 S (m_{f} + 1) + \vartheta_{0})\frac{1}{\tau_{\gamma}\tau_{\eta}} - \vartheta_{0}\frac{1}{\tau_{\gamma}^{2}} )t_{e}\\
    &= - 2 S \cdot m_{f}\frac{1}{\tau_{\eta}} - 2 \vartheta_{0}\frac{1}{\tau_{\gamma}},\\
    (- S \cdot m_{f}\tau_{\gamma}^{2} &- (2 S (m_{f} + 1) + \vartheta_{0})\tau_{\gamma}\tau_{\eta} - \vartheta_{0}\tau_{\eta}^{2} )t_{e}\\
    &= - 2 S \cdot m_{f}\tau_{\gamma}^{2}\tau_{\eta} - 2 \vartheta_{0}\tau_{\gamma}\tau_{\eta}^{2}.
\end{align*}
This leads to our expression for $t_{e}:$
\begin{align*}
    t_{e} = \frac{2 \tau_{\gamma} \tau_{\eta}( S \cdot m_{f}\tau_{\gamma} + \vartheta_{0}\tau_{\eta})}{S \cdot \tau_{\gamma} (m_{f}\tau_{\gamma} + 2 (m_{f} + 1)\tau_{\eta}) + \vartheta_{0} \tau_{\gamma} \tau_{\eta} + \vartheta_{0}\tau_{\eta}^{2}}.
\end{align*}
We now insert this expression in Equation \ref{eq:Is} and get:
\begin{align*}
    I_{l}(S) = \frac{h}{e^{\frac{t_{e}}{\tau_{\eta}}} - 1} = \frac{h}{\exp \left( \frac{2 \tau_{\gamma} ( S \cdot m_{f}\tau_{\gamma} + \vartheta_{0}\tau_{\eta})}{S \cdot \tau_{\gamma} (m_{f}\tau_{\gamma} + 2 (m_{f} + 1)\tau_{\eta}) + \vartheta_{0} \tau_{\gamma} \tau_{\eta} + \vartheta_{0}\tau_{\eta}^{2}} \right) - 1}.
\end{align*}
To make sure that our activation function $f(S)$ is $0$ at $S = \vartheta_{0}/2$ we choose our activation function to be:
\begin{align}
\label{eq:approxactivfun_appendix}
    f(S) &= I_{l}(S) - I_{l}(\frac{\vartheta_{0}}{2}) = \frac{h}{\exp\left( \frac{2 m_{f} \tau_{\gamma}^{2} S + 2 \vartheta_{0}\tau_{\eta} \tau_{\gamma}}{ \tau_{\gamma} (m_{f}\tau_{\gamma} + 2 (m_{f} + 1)\tau_{\eta}) S + \vartheta_{0} \tau_{\gamma} \tau_{\eta} + \vartheta_{0}\tau_{\eta}^{2}} \right) - 1} - c,
\end{align}
for $S > \frac{\vartheta_{0}}{2}$ and $f(S) = 0$ for $S <= \frac{\vartheta_{0}}{2}$ with $c = I_{l}(\frac{\vartheta_{0}}{2})$.\\

\renewcommand{\thetable}{SI\arabic{table}}
\newpage
\section*{Parameters used in the Arousal attention method}

\begin{table}[H]
  \caption{Parameters used in for the Arousal attention method.}
  \centering
  \vspace{0.2cm}
  \begin{tabular}{lrrrrrr}

 DataSet & Selectivity(\%) & $t_{sa}$(ms) & $\vartheta_{0-lp}$ & $\vartheta_{0-hp}$ & $p_{1}$ & $p_{2}$  \\

    \midrule 
    IRIS    & $32.00$ & $150$ & $0.80$ &  $0.17$ & $2.1864$ & $-342.2727$\\
    SONAR   & $44.23$ & $150$ & $1.30$ &  $0.35$ & $3.7591$ & $-525.0000$\\
    MNIST & $ 1.13$ & $200$ & $0.60$ &  $0.12$ & $2.1121$ & $-405.7576$\\
    CIFAR-10& $48.07$ & $250$ & $0.15$ &  $0.05$ & $7.7262$ & $-1.87e^{+03}$\\
    CIFAR-100& $70.50$ & $350$ & $0.10$ &  $0.03$ & $7.0000$ & $-2.42e^{+03}$\\
    LSVRC-2012& $ 76.64 $ & $350$ & $0.15$ &  $0.05$ & $2.5536$ & $-895.1786$\\

  \end{tabular}
\end{table}

\section*{Supplementary results}
\renewcommand{\thefigure}{SI\arabic{figure}}

\begin{figure}[H]
\centering
  \includegraphics[width=\textwidth]{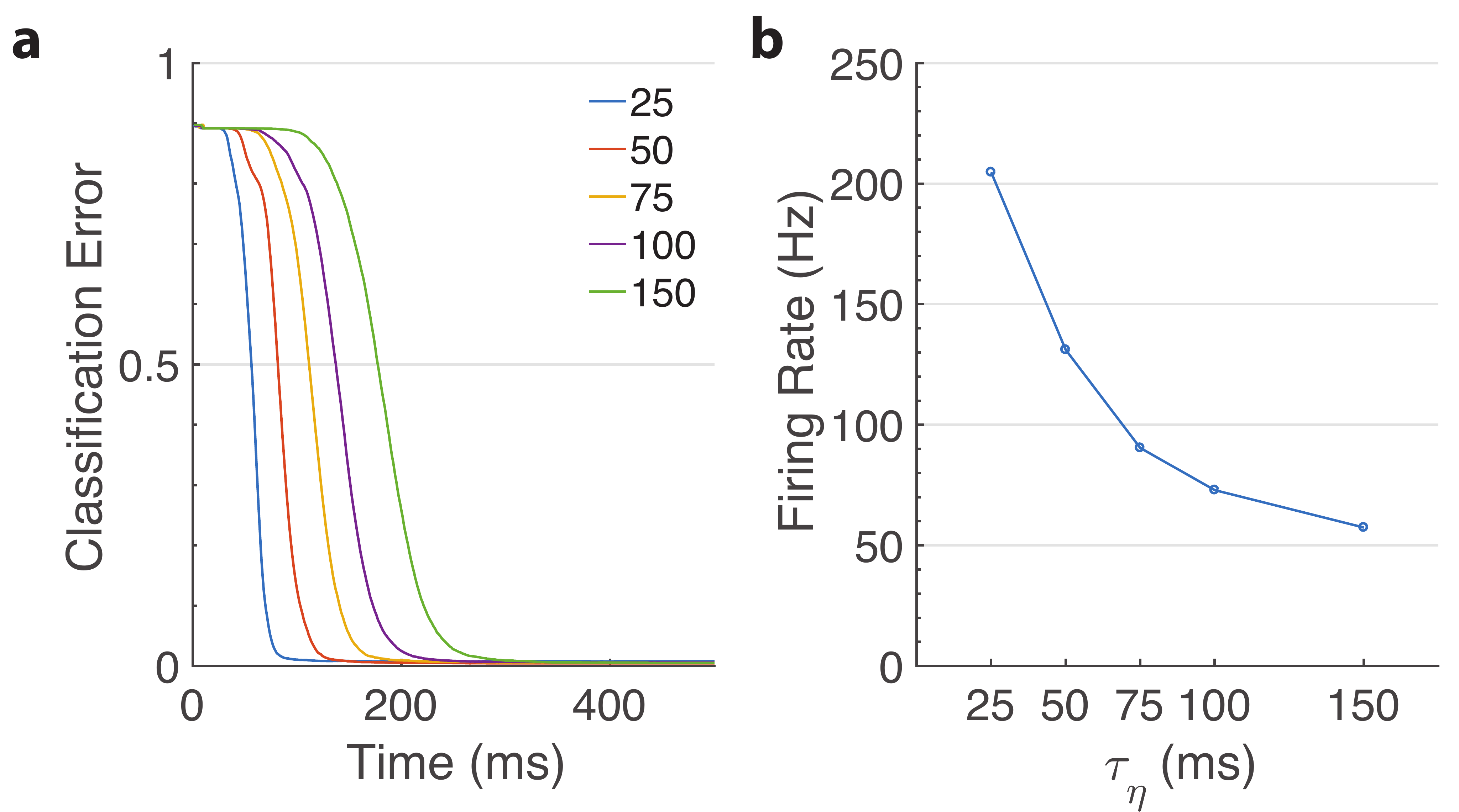}
  \caption{Effects of $\tau_{\eta}$ on MNIST. {\bf a.} The classification error over time is shown for increasing values of $\tau_{\eta}$s, $[25, 50, 75, 100, 150]$ms. Note that changing $\tau_{\eta}$ changes the transfer function shape, and thus different networks were trained. The plotted results are obtained with $\vartheta_0=0.05$. MT visibly increases for longer $\tau_{\eta}$s. {\bf b.} Networks' firing rates. Longer $\tau_{\eta}$s require less spikes to approximate a signal.  }
  \label{fig:figSI1}
\end{figure}

\begin{figure}[H]
\centering
  \includegraphics[width=\textwidth]{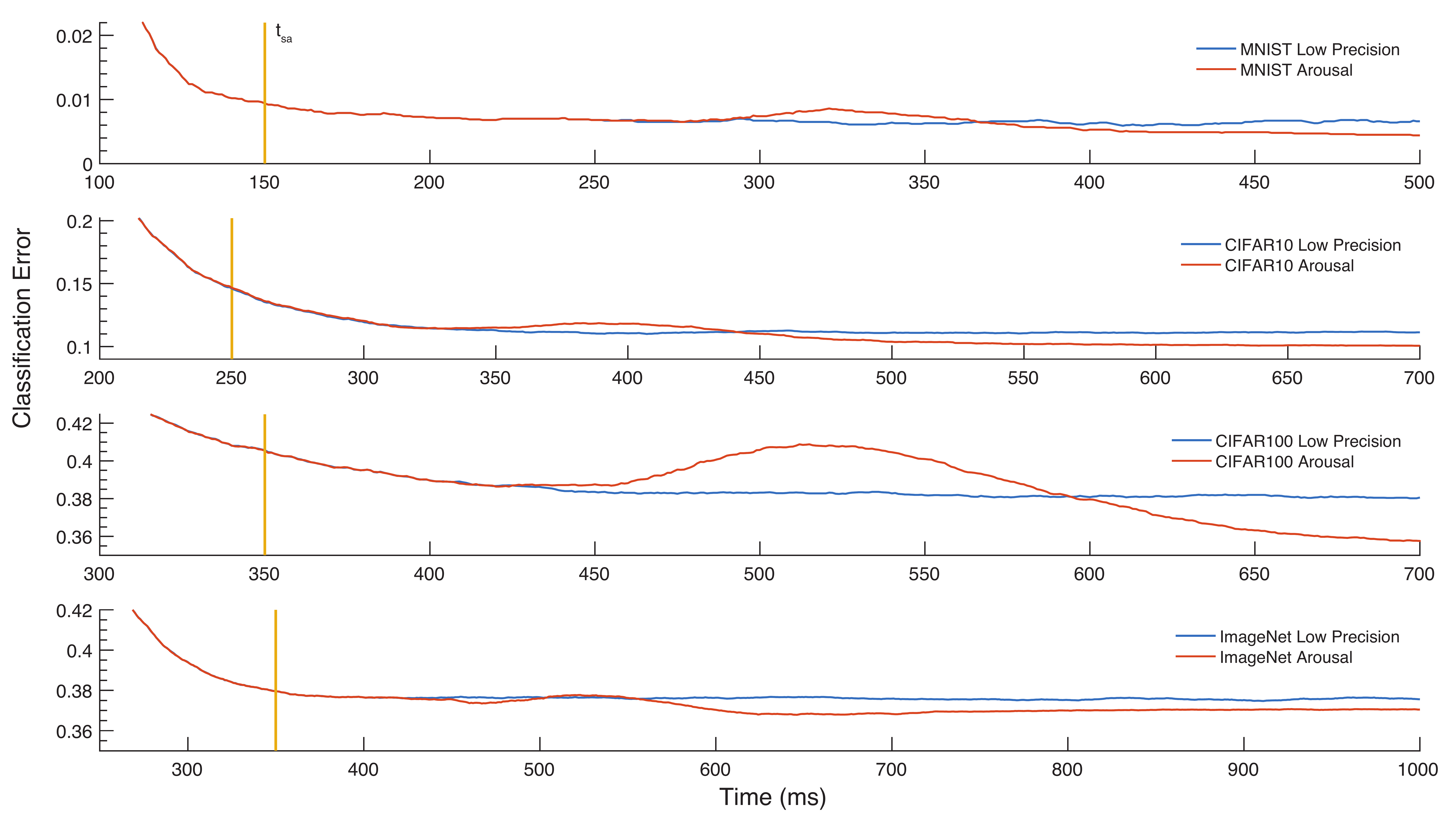}
  \caption{Classification error over time. The effect of the Arousal method on the classification error is reported for MNIST, CIFAR-10, CIFAR-100, and the Imagenet LSVRC-2012. The vertical line denotes the moment in time, $t_{sa}$, where the outputs start being accumulated. Selection for Arousal is then determined 50ms later. The increase of the firing rate on selected images causes a brief loss of accuracy, after which a lower classification error is reached. }
  \label{fig:figSI2}
\end{figure}

\newpage
\bibliography{SNN}

\subsection*{Acknowledgement}

DZ is supported by NWO project 656.000.005.

\end{document}